\documentclass[11pt]{article}

\usepackage[final]{acl}

\usepackage{times}
\usepackage{latexsym}

\usepackage[T1]{fontenc}

\usepackage[utf8]{inputenc}

\usepackage{microtype}

\usepackage{graphicx}

\usepackage{booktabs}

\usepackage{amsmath}

\usepackage{float}

\usepackage[most]{tcolorbox}
\usepackage{xcolor}

\newtcolorbox{personabox}[1]{
  enhanced,
  breakable,
  colback=gray!4,
  colframe=gray!45,
  boxrule=0.4pt,
  arc=2pt,
  left=8pt,
  right=8pt,
  top=6pt,
  bottom=6pt,
  title=\textbf{#1},
  coltitle=black,
  colbacktitle=gray!15,
  fonttitle=\normalsize,
  before skip=8pt,
  after skip=8pt
}

\title{Creating Multilingual Mental Health Dialogue Datasets: Limits of Persona-Based Localization via Nationality and Language}

\newcommand*{\marked}[1]{{#1}}

\author{
Yunkai Xu \and Saeed Abdullah \\
Pennsylvania State University, United States \\
\texttt{\{yunkai, saeed\}@psu.edu}
}

\begin{document}
\maketitle
\begin{abstract}
%
AI and large language models (LLMs) have emerged as promising tools to address global mental health challenges. Despite the global nature of these challenges, there remains a critical shortage of high-quality datasets for training and evaluating such systems. To mitigate this gap, researchers increasingly generate synthetic clinical personas to simulate user data and test digital mental health support systems. However, most validated personas rely on English-centric contexts. This paper investigates whether similar persona-based methods can be used to generate multilingual mental health datasets. We modified nationality and language parameters in personas to generate clinical dialogues in Mandarin, Bengali, and Hindi. We then examined how different LLMs perform when evaluating the depression severity of these generated multilingual datasets against the baseline in English. Our findings indicate that just adding nationality and language parameters in personas might not be adequate, as it can introduce clinical inconsistency across languages. LLM judge models often exhibit inaccuracies in assessing depression severity in non-English texts, with performance varying across different models. This exposes the systemic limitations of applying English-centric personas to multilingual contexts. Ultimately, our work highlights the urgent need for culturally responsive data generation to ensure equitable mental health systems globally.


All personas and the full pipeline are publicly available at \textbf{\href{https://github.com/Xuyk021/CLPsych2026workshop.git}{GitHub link}}.
\end{abstract}

\section{Introduction}
Depression presents one of the significant global challenges among mental health disorders~\cite{moitraGlobalMentalHealth2023a,liuTemporalSpatialTrend2024}. Regions with limited medical infrastructure, such as Bangladesh, experience a severe shortage of mental health professionals~\cite{hasanCurrentStateMental2021}. Artificial intelligence (AI) and large language models (LLMs) provide a potential medium for depression screening and continuous support~\cite{chen2024depression,ignashina2025llm,li2025customizable,zhang2025generative}. However, the benefits of these technological advancements are also not distributed equitably across regions. Prior work has primarily focused on English~\cite{zhang2025generative} and emerging evidence suggests that language itself plays a critical role in shaping LLM performance, with more pronounced effects observed in non-English languages~\cite{jin2023better, raihan2026large}. This marginalizes populations that use other languages, creating systemic biases and representation disparities in digital mental health assessments.

Recently, researchers have increasingly utilized synthetic data to mitigate these biases in underrepresented languages. This need is particularly pronounced in the mental health domain, where access to real-world data is further constrained by privacy concerns~\cite{kang2024synthetic, lorge2025detecting}.
Among various approaches to synthetic data generation, persona-based methods have emerged as an effective strategy, particularly for scaling up data generation~\cite{zhang2018personalizing,ge2025scaling,jandaghi2024faithful}. In the mental health field, some personas simulate specific psychological states, such as those defined by the Beck Depression Inventory II (BDI-II)~\cite{weinerBDI2010}, acting as standardized patients for training and evaluating clinical dialogue systems~\cite{wangTalkDepClinicallyGrounded2025}.


However, the majority of validated clinical synthetic personas originate from English and Western contexts~\cite{wangTalkDepClinicallyGrounded2025}. It might be feasible to address the current language gap by modifying specific demographic and linguistic variables within the original English persona prompt. However, prior studies demonstrate that altering a single persona parameter affects downstream behavior in other domains~\cite{weeberOnePersonaMany2026, kamruzzaman2025anger}. It remains unclear whether this parameter-based localization preserves the original clinical symptoms when generating interactions in non-English contexts.

This study investigates the preservation of clinical features following the parameter-based adaptation of synthetic personas. We designed a controlled experiment using a clinically validated English persona as a baseline~\cite{wangTalkDepClinicallyGrounded2025}. By modifying only the language and nationality variables, we generated parallel personas for different regions. We then used independent judges implemented with different LLMs to assess the depression severity reflected in the generated chat histories. Our study addresses the following research questions:
\begin{itemize}
    \item \textbf{RQ1:} How does parameter-based localization affect the clinical consistency of symptoms expressed by synthetic personas across different languages?
    \item \textbf{RQ2:} How do diverse LLMs model judges vary in their capability and certainty when assessing depression severity in non-English synthetic conversations generated by the personas?
\end{itemize}

By addressing these questions, our research provides the following key contributions:
\begin{itemize}
    \item We provide empirical evidence that modifying persona parameters leads to disparities in depression level representation and weakens alignment with clinical severity levels in dialogue datasets generated in Mandarin, Bengali, and Hindi.
    \item We highlight the systemic limitations of applying synthetic personas to multilingual artificial intelligence systems, and argue for treating them as clinical artifacts. Specifically, we call for rigorous output-level validation and language-specific evaluation to maintain cross-lingual consistency.
\end{itemize}

\section{Related Works}
\subsection{Multilingual Large Language Models in Mental Health}

Mental health care resources remain unevenly distributed worldwide, with many regions lacking trained clinicians and accessible services~\cite{hasanCurrentStateMental2021}. Recent work has therefore started to explore how LLMs might support mental health care in non-English contexts~\cite{zhang2025generative}. Existing efforts span a range of linguistic and cultural settings. In Chinese-speaking contexts, prior studies have developed LLM-based support for cognitive behavioral therapy and deployed chatbot systems for anxiety and depression support~\cite{na2024cbtllm,chen2025comparison}. Similar efforts are beginning to emerge in South Asia. For instance, one study considered the potential of LLMs for suicide prevention in India~\cite{chakraborty2025promise}, while another proposed an LLM-supported intervention for postpartum mental health among women in Bangladesh~\cite{ahmed2025conceptual}. Taken together, these studies point to growing interest in multilingual mental health support, but they also remain focused on particular languages and application scenarios.

In principle, fair multilingual models should maintain consistent performance across languages. However, some studies report clear performance gaps between high-resource languages such as English and lower-resource languages such as Bengali~\cite{bhowmik2025evaluating}. A main reason for this gap is the imbalance in training and fine tuning data. High-quality clinical data are common in English datasets but rare in many other languages. This shortage limits the use of language models for mental health support in different cultural settings.

\subsection{Datasets for Depression Detection and Synthetic Data}

Existing depression detection datasets are primarily English-centric, sourced from platforms like Reddit and X~\cite{gui2019cooperative, rissola2020dataset, naseem2022early, parapar2026overview}. While resources in languages such as Chinese, Japanese and Portuguese have emerged~\cite{santos2024setembrobr, xiao2026jiraibench, agarwal2021deep}, they remain limited to single-language settings. As a result, they do not address the broader challenge of capturing linguistic and cultural diversity across multiple languages. 

Collecting large-scale clinical data in non-English context is further hindered by privacy regulations and the scarcity of online mental health communities~\cite{kang2024synthetic, lorge2025detecting}. To bridge this gap, researchers often translate English datasets into other languages~\cite{yang2019pawsx, myung2024blend, jin2023better}. However, human translation is costly and difficult to scale, while machine translation often fails to capture cultural nuances or specific mental health expressions~\cite{bhowmik2025evaluating, raihan2026large}.

To address data scarcity and ethical constraints, recent studies leverage LLMs for synthetic data generation~\cite{wang2024survey}. In mental health, this includes simulating conversational therapy sessions~\cite{zhezherau2024hybrid}, patient needs~\cite{ronan2025langchainbased}, and patients' clinical synopses to balance severity distributions~\cite{kang2024synthetic}. Others have generated context-enhanced social media data to identify psychosocial risks~\cite{garg2026leveraging}. Despite these advances, existing work focuses almost exclusively on single-language generation. However, existing work often focuses on generating data within a single language. It remains unclear how synthetic representations generalize across languages and whether they can preserve clinically meaningful signals in a multilingual framework.



\subsection{Personas as a Source of Synthetic Data}
Personas represent archetypal people through fictional yet data-informed profiles, enabling designers to reason about the characteristics and goals of a target population~\cite{pruitt2003personas}.
LLMs can generate high-quality personas comparable to human experts~\cite{schuller2024generating}, making them a practical alternative when traditional data collection is constrained~\cite{salminen2025use}.
LLMs have further enabled data-driven persona development \cite{jung2025personacraft}, maintaining high fidelity even in specialized domains \cite{kaur2025synthetic}.

In mental health, personas are increasingly used to synthesize therapeutic dialogue data to bypass data scarcity. For instance, \citet{jandaghi2024faithful} proposed a generator-critic architecture for faithful persona-based interactions, while \citet{wu2025personas} leveraged persona traits to modulate emotional support sessions. To ensure clinical grounding, \citet{wangTalkDepClinicallyGrounded2025} incorporated specific attributes like BDI-II scores into simulations. To standardize these efforts, the PatientHub framework \cite{sabour2026patienthub} provides a modular system for reproducible simulated patient deployment.


However, the use of synthetic personas introduces risks associated with data validity and algorithmic bias~\cite{li2025llm, batzner2025whose}. Variations in the components of a persona can cause inconsistencies in downstream performance~\cite{wu2025personas}. \citet{kamruzzaman2025anger} identified that assigning nationality-specific personas to LLMs results in emotional stereotypes, as the models disproportionately attribute certain emotions to specific countries. Furthermore, assigned nationality personas can alter how models perceive different nations, which often leads to more favorable treatment of Western regions compared to others~\cite{kamruzzaman2025exploring}.

While prior work highlights cultural bias in LLMs in healthcare, the majority of validated clinical personas are still developed within English-speaking contexts. There remains a lack of evidence concerning whether the simple modification of nationality and language parameters within English-centric prompts is sufficient to generate valid clinical dialogues in other languages. Our work addresses this gap by investigating how such adaptations influence the clinical accuracy of depression assessments across different linguistic settings.

\section{Method}
\begin{figure*}[!t]
    \centering
    \includegraphics[width=\textwidth]{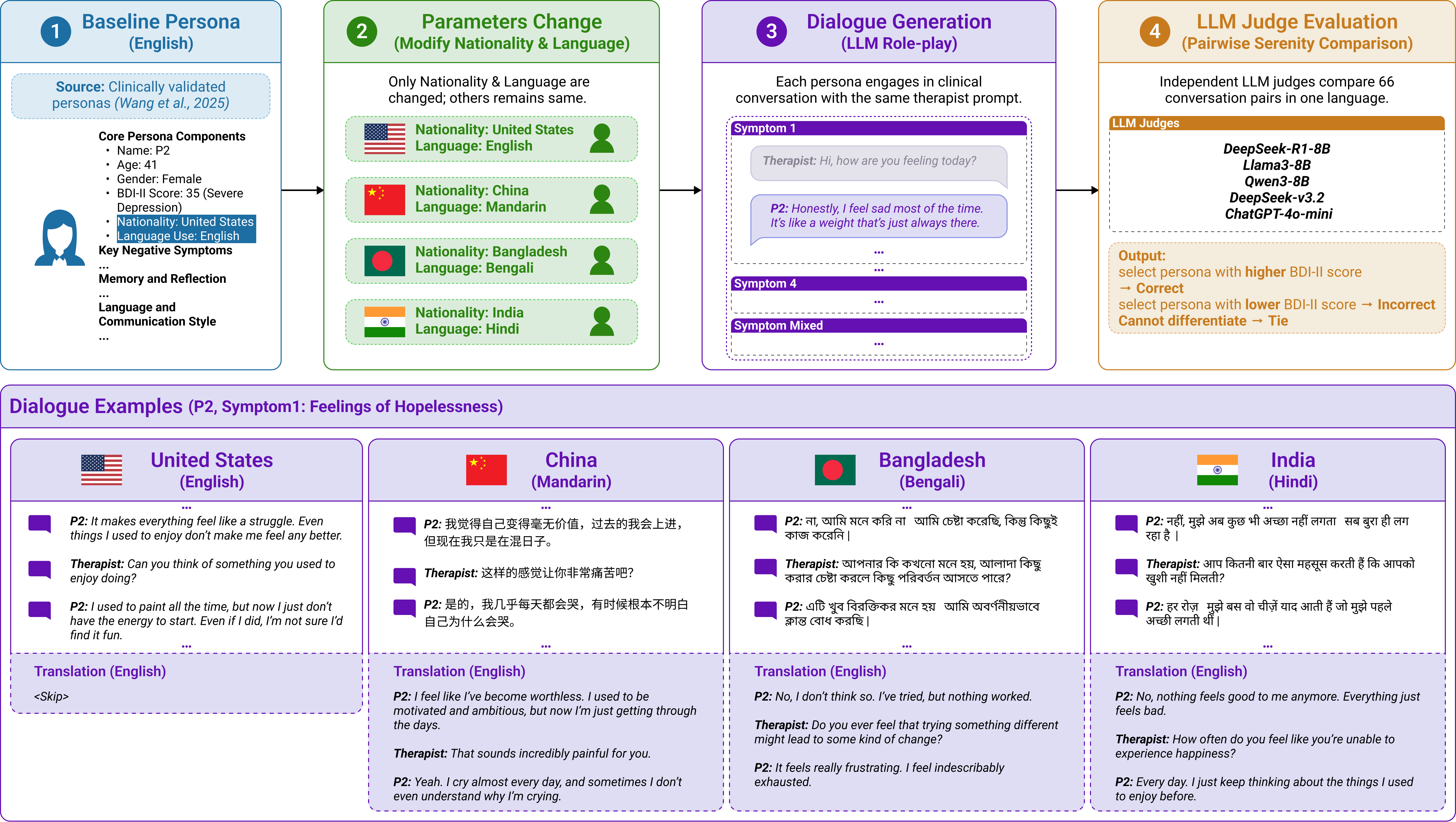}
    \caption{
    Overview of the multilingual synthetic dialogue generation and evaluation workflow. 
    Example dialogue sessions are translated into English for readability using Google Translate\protect\footnotemark.
    }
    \label{fig:workflow}
\end{figure*}

\subsection{Persona Construction}


\marked{We adapt the workflow illustrated in Figure~\ref{fig:workflow} and the clinically validated baseline personas from prior work by Wang et al.~\cite{wangTalkDepClinicallyGrounded2025}.} The personas consist of several dimensions that capture demographic attributes, depression symptoms, communication patterns, and social context \marked{(see Appendix~\ref{app:persona_prompt})}. Table~\ref{tab:persona_dimensions} summarizes these dimensions and their sources.
These elements serve specific functions in clinical depression diagnosis. For instance, a documented life history is an important component in understanding depression and related mental health outcomes, and analyzing detailed life histories helps identify pathways that link individual experiences with mental health trajectories~\cite{singerLinkingLifeHistories1998}. Therefore, we include life history information when constructing depression personas. In total, we use 12 personas (Table~\ref{tab:personas}). All personas include the dimensions described above, and the content in these dimensions differ across personas. We focus on personas' depression characteristics by varying symptoms, severity levels, and BDI-II scores to represent different levels of depression.

\begin{table*}[t]
\centering
\caption{Dimensions used to construct clinical personas.}
\label{tab:persona_dimensions}

\begin{tabular}{p{0.4\textwidth}p{0.54\textwidth}}
\toprule
\textbf{Dimension} & \textbf{Sources} \\
\midrule
Age & \cite{faravelliGenderDifferencesDepression2013} \\

Gender & \cite{faravelliGenderDifferencesDepression2013,enselRoleAgeRelationship1982} \\

BDI-II score & \cite{weinerBDI2010} \\

Depression symptoms and severity & \cite{DSMfive2013} \\

Communication style & \cite{smirnovaLanguagePatternsDiscriminate2018,almosaiwiAbsoluteStateElevated2018} \\

Life history & \cite{singerLinkingLifeHistories1998} \\

Social context & \cite{aseltineDepressionSocialDevelopmental1994,brownSocialRolesContext2002} \\

Clinician behavior constraints & \cite{bickleyBatesGuidePhysical2012} \\

Social media use & \cite{huangMetaanalysisProblematicSocial2022,berrymanSocialMediaUse2018} \\
\bottomrule
\end{tabular}
\end{table*}

We introduce two new variables to these baseline personas: \textbf{Nationality} and \textbf{Language Use}. Many prior studies show that Nationality can change how a persona behaves in downstream tasks~\cite{kamruzzaman2025exploring,kamruzzaman2025anger}. We include the Language Use variable to dictate the output language of the generated text. We represent specific geographic regions by pairing nationality with language use (United States-English, Chinese-Mandarin, Bangladeshi-Bengali, and Indian-Hindi). 
We focused on these regions and languages as they are among the most widely spoken languages globally \cite{ethnologue2023}.
The remainder of the persona prompt remains in English and keeps the same content and structure as the original clinically validated persona. Using this procedure, we generate distinct personas for different target regions, with 12 personas in each language and 48 personas in total.

This approach follows prior work that adopts controlled modification to enable systematic comparison across persona settings~\cite{sakai2025somatic}.

\footnotetext{\url{https://translate.google.com/}}

\subsection{Evaluation: Pairwise Severity Comparison}
\label{sec:evaluation}

To validate the clinical consistency of our generated personas, we mainly replicate a pairwise depression severity differentiation task~\cite{wangTalkDepClinicallyGrounded2025} and extend this to a multilingual context to test our personas.
The evaluation process consists of two stages:
\begin{enumerate}
    \item \textbf{Dialogue Generation:} We employ an independent LLM-based therapist agent (powered by ChatGPT-4o-mini) to conduct semi-structured clinical interviews with each persona. \marked{The agent using the simple prompt was adapted from \cite{wangTalkDepClinicallyGrounded2025}, with additional instructions added to ensure that the generated responses remained consistent with the language settings specified in the persona (see Appendix~\ref{app:therapist_prompt}).} For each persona, this interaction generated five distinct dialogue \marked{sessions}. \marked{Specifically, four sessions were designed to focus on individual target symptoms defined in the persona, while an additional mixed-symptom session combined all four symptoms. The dialogue session consisted of five to seven turns.} Unlike \citet{wangTalkDepClinicallyGrounded2025}'s work, our personas possess specific nationalities, and the therapist interacts with them in their \textbf{corresponding languages}. This results in a multilingual corpus consisting of 60 dialogue sessions per language.
    \item \textbf{Blind Pairwise Judging:} An independent LLM-based judge is presented with generated dialogues from two personas with different depression severity levels (based on BDI-II scores). The judge is blinded to the underlying persona configurations and scores. Its task is to determine which patient exhibits a higher level of depression severity, or if they are indistinguishable (i.e., Persona A > Persona B, B > A, or "Tie" as a tie case). Consequently, each evaluation cycle comprises 66 pairwise comparisons \marked{($N_{\text{total}} = \binom{12}{2} = 66$)}.
\end{enumerate}

This setup evaluates whether the clinical cues of depression remain detectable and consistent when expressed through different languages and cultural backgrounds, and whether the LLM judge can maintain diagnostic sensitivity despite the changes in the personas.

\subsection{Evaluation Metrics}

Replicated from ~\citet{wangTalkDepClinicallyGrounded2025}'s work, We also evaluate model performance using three metrics: overall accuracy, same-level error rate, and tie distance.

\textbf{Overall Accuracy.} 
This metric evaluates the LLM judge's proficiency in correctly identifying the relative depression severity within each pair. A prediction is counted as correct if the judge successfully selects the dialogues associated with the persona that possesses the higher ground-truth BDI-II score. \marked{Cases where the model outputs selects the persona with the lower score are treated as incorrect. Tie responses were treated as a separate outcome category. Thus, the total number of comparisons can be expressed as:}

\begin{equation}
\label{eq:total_comparisons}
N_{\text{total}} =
N_{\text{correct}} + N_{\text{incorrect}} + N_{\text{tie}}
\end{equation}
\marked{Overall accuracy is then computed over non-tie comparisons:}
\begin{equation}
\label{eq:overall_accuracy}
\text{Overall Accuracy} =
\frac{N_{\text{correct}}}{N_{\text{total}} - N_{\text{tie}}} \times 100\%
\end{equation}
\textbf{Same-Level Error Rate.}
This metric analyzes the quality of model errors. 
When a model makes an incorrect prediction, we examine whether the two personas belong to the same BDI-II severity category. 
We define Same-Level Error Rate as the percentage of incorrect predictions where the two evaluated personas actually belong to the same severity level. 
This metric captures errors that occur near clinical severity boundaries.

\begin{equation}
\label{eq:same_level_error}
\text{Same-Level Error Rate} =
\frac{E_{\text{same}}}{E_{\text{total}}} \times 100\%
\end{equation}

where $E_{\text{same}}$ is the number of errors between personas within the same severity category, and $E_{\text{total}}$ is the total number of incorrect predictions.
Errors that cross severity boundaries indicate larger clinical misjudgments.

\textbf{Tie Distance.}
Models occasionally refuse to select one chat history and instead output a tie. 
We measure the Average Tie Distance to quantify the severity gap between the personas in these tie cases. 
The distance is the absolute difference between the underlying BDI-II scores of the two personas. 
A larger tie distance indicates that the model expresses uncertainty even when the actual severity difference is large.

\subsection{Model Use}





We use several LLMs as judges, including proprietary models (GPT-4o-mini \cite{openaiGPT4TechnicalReport2024}, DeepSeek-V3.2 \cite{deepseekaiDeepSeekV32PushingFrontier2025}) and open-weight models (LLaMA3.1-8B \cite{grattafioriLlama3Herd2024}, Qwen3-8B \cite{yangQwen3TechnicalReport2025}, and DeepSeek-R1-8B \cite{deepseekaiDeepSeekR1IncentivizingReasoning2025}). We use the 8B versions to align with prior work \cite{wangTalkDepClinicallyGrounded2025} and maintain efficiency.

In the evaluation task, each model compares the dialogues from two personas to decide which one shows more severe depression. The models do not see the persona prompts, so the assessment is based only on the generated text.

\section{Findings}

\subsection{Cross-Model and Cross-Lingual Results}
\subsubsection{Overall Accuracy and Cross-Lingual Disparity}

\marked{We present the overall accuracy of the evaluated models across four languages in Table~\ref{tab:overall_accuracy}.} The results reveal a persistent English advantage across all language models. Performance in English consistently serves as the upper bound for each model. 

ChatGPT-4o-mini and DeepSeek-v3.2 achieve the highest overall accuracy. ChatGPT-4o-mini maintains the lowest cross-lingual variability with a standard deviation of 4.17. DeepSeek-v3.2 is slightly higher than ChatGPT-4o-mini. Both of them demonstrate relatively robust calibration in interpreting depression severity across different linguistic contexts. 

Smaller open-weight models demonstrate substantial performance drops when processing non-English personas. DeepSeek-R1-8B and Llama3-8B exhibit high performance fluctuations across languages, with standard deviations of 16.15 and 16.69 respectively. DeepSeek-R1-8B achieves 98.21\% accuracy in English but drops to 63.33\% in Bengali and 67.24\% in Hindi. This indicates that mental disorder severity interpretation capabilities in smaller models remain heavily biased toward English contexts.

\begin{table}
\centering
\caption{Overall accuracy across models and languages.}
\label{tab:overall_accuracy}
\resizebox{\columnwidth}{!}{
\begin{tabular}{lrrrr}
\hline
\textbf{Model} & \textbf{Bengali} & \textbf{English} & \textbf{Hindi} & \textbf{Mandarin} \\
\hline
ChatGPT-4o-mini & 92.42 & 95.45 & 87.88 & 86.36 \\
DeepSeek-v3.2 & 90.91 & 93.75 & 83.33 & 84.85 \\
DeepSeek-R1-8B & 63.33 & 98.21 & 67.24 & 84.62 \\
Llama3-8B & 45.45 & 84.13 & 65.38 & 75.76 \\
Qwen3-8B & 71.21 & 89.39 & 73.44 & 80.30 \\
\hline
\end{tabular}
}
\end{table}

\subsubsection{Same-Level Error Rate and Cross-Lingual Disparity}

\begin{table}[t]
\centering
\caption{Same-level error rates (\%) across languages and models. Higher values indicate fewer cross-severity errors. Bold values indicate the highest score within each language.}
\label{tab:error_metrics}

\resizebox{\columnwidth}{!}{
\begin{tabular}{lcccc}
\hline
\textbf{Model} 
& \textbf{Bengali} 
& \textbf{Hindi} 
& \textbf{Mandarin} 
& \textbf{English} \\
\hline
ChatGPT-4o-mini & 80.00 & \textbf{87.50} & 66.67 & 66.67 \\
DeepSeek-v3.2   & \textbf{100.00} & 72.73 & 70.00 & \textbf{100.00} \\
DeepSeek-R1-8B  & 22.73 & 31.58 & \textbf{80.00} & \textbf{100.00} \\
Llama3-8B       & 23.33 & 33.33 & 50.00 & 60.00 \\
Qwen3-8B        & 31.58 & 29.41 & 61.54 & 85.71 \\
\hline
\end{tabular}
}

\end{table}

Cross-language differences manifest in overall accuracy and in the nature of the errors. We analyze whether incorrect model answers stay within the same depression severity level or cross-severity boundaries. Table~\ref{tab:error_metrics} details these error distributions.

Models exhibit severe cross-lingual inconsistency in their error calibration. English consistently corresponds to fewer cross-severity mistakes for most models. For example, DeepSeek-R1-8B maintains a 100\% Same-Level Error Rate in English. This means all its English mistakes occur between personas of identical severity tiers. This is important because many studies are concerned with severity categories rather than exact scores~\cite{naseem2022early, peng2026framework}, making within-tier errors less problematic than cross-tier ones. However, its Same-Level Error Rate drops to 22.73\% in Bengali. This indicates that Bengali errors frequently cross-severity boundaries. 

Llama3-8B and Qwen3-8B show similar patterns. Llama3-8B confines 60.00\% of its English errors to the same severity level, compared to 23.33\% in Bengali and 33.33\% in Hindi. DeepSeek-v3.2 is an exception. It maintains consistently high Same-Level Error Rates across all languages, indicating a robust calibration in distinguishing adjacent severity levels regardless of the input language. These findings demonstrate that language changes affect both the performance level and the structural consistency of clinical severity judgments.

\subsubsection{Uncertainty and Tie Behavior}

We further analyze model uncertainty by examining explicit tie outputs. Table~\ref{tab:tie_cases} summarizes the frequency and severity distance of tie cases. 

Tie behavior varies unevenly across models and languages. DeepSeek-R1-8B produces the highest number of ties across all four languages, totaling 25 cases. Llama3-8B produces 14 ties, but these are heavily concentrated in Bengali and Hindi. ChatGPT-4o-mini never outputs a tie, and DeepSeek-v3.2 rarely produces ties.

The context of these ties also differs significantly between models. DeepSeek-v3.2 expresses uncertainty exclusively in near boundary cases, with a mean tie distance of 4.00. DeepSeek-R1-8B and Llama3-8B frequently express uncertainty even when severity differences are large. Their mean tie distances are 10.80 and 14.14 respectively. 

Furthermore, the difficulty threshold that triggers a tie depends on the language. For DeepSeek-R1-8B, the average tie distance is 15.12 in Hindi compared to 7.33 in Bengali and 8.90 in English. This indicates that the severity gap required to confuse the model is not constant and shifts substantially according to the linguistic context.

\begin{table}
\centering
\caption{Overall tie counts and average tie distance by model.}
\label{tab:tie_cases}
\resizebox{\columnwidth}{!}{
\begin{tabular}{lrrr}
\hline
\textbf{Model} & \textbf{Tie Count} & \textbf{Avg Distance} & \textbf{Max Distance} \\
\hline
DeepSeek-R1-8B & 25 & 10.80 & 34 \\
Llama3-8B & 14 & 14.14 & 28 \\
Qwen3-8B & 2 & 5.50 & 6 \\
DeepSeek-v3.2 & 2 & 4.00 & 5 \\
ChatGPT-4o-mini & 0 & - & - \\
\hline
\end{tabular}
}
\end{table}

\subsection{Persona-based Methods for Multilingual Dataset}

The high accuracy and low cross boundary error rates observed in English evaluations suggest that the baseline English personas successfully generate distinct and recognizable clinical traits, which is consistent with prior studies~\cite{wangTalkDepClinicallyGrounded2025}. However, the decline in judge accuracy for Bengali and Hindi indicates a degradation in the underlying text. The increased frequency of cross-severity errors and large distance tie cases in non-English languages shows that the generated dialogues may lack clear clinical signals in the original personas. 

Overall, these results collectively demonstrate that modifying the nationality and language labels within an English-centric persona is insufficient to create viable multilingual clinical personas. The shallow localization process introduces clinical ambiguity and this problem is much more visible in smaller 8B models, causing the resulting text to lose alignment with the intended BDI-II severity scales.



\section{Discussion}

\subsection{Why Minimal Localization in Personas Fails to Preserve Clinical Cues}
\label{sec:discussion_51}
This study examined a strategy for multilingual persona construction: starting from an expert-validated English persona prompt and replacing only nationality and language to generate non-English personas. We then evaluated the resulting dialogues through blinded severity judgment. Across LLM judge models, English consistently acted as the upper bound, while non-English dialogues more often led to severity misclassification and ties. This pattern suggests that simple parameter substitution does not reliably preserve clinically relevant cues across languages.

\marked{One potential reason is that changing nationality and language also alters the cultural context embedded in the persona, while symptom expression may vary substantially across cultures and therefore may not be preserved consistently~\cite{goodmann2021factor, jovanovic2026depression,bradshaw2026demographic}.} The same BDI-II level can surface in different ways depending on the conversation, as the expression of and response to depression vary significantly across cultures~\cite{teja1971depression}. This issue has been observed not only in real clinical interactions but also in synthetic sessions generated from personas~\cite{sakai2025somatic}. When these expressive patterns shift, the resulting dialogue may still describe similar experiences but fail to present them in a form that judges can consistently interpret. Prior work reports similar cross-lingual gaps in health-related tasks and shows that models often perform better when queries are expressed in English~\cite{jin2023better}. Therefore, our findings extend this observation to persona-driven dialogue generation.

\marked{In addition, both nationality and language do not act as neutral parameters for the LLM dialogue generation and the LLM judges.} Evidence from cultural benchmarks shows that model performance can vary across regions and languages on everyday knowledge tasks, and that performance can remain higher in English than in the local language for some low-resource cultures~\cite{myung2024blend}. Furthermore, assigning nationality has been shown to change model outputs in systematic ways, including how emotions and social attributes are expressed~\cite{kamruzzaman2025exploring,kamruzzaman2025anger}. In our setting, this means that the generated dialogue may introduce additional culture-specific interaction patterns that were not present in the original English persona. As a result, symptom descriptions can shift in structure and emphasis, which reduces the clarity of severity signals even when the underlying prompt remains unchanged.

\subsection{Implications for Multilingual Mental Health Persona Construction and Evaluation}

The findings have direct implications for the use of personas in multilingual mental health data generation and evaluation. First, multilingual dialogues produced through simple parameter replacement should not be treated as equivalent samples across languages. In our study, non-English dialogues show not only lower accuracy but also more cross-severity errors and unstable tie behavior. This indicates a shift in the underlying data distribution. Similar effects have been reported in multilingual mental health benchmarks, where model performance drops when tasks rely on translated rather than native data, with variation across languages and translation quality~\cite{raihan2026large}.

Second, persona construction requires stronger methodological control in clinical settings. Prior work has identified common issues in persona-based research, including weak specification of target populations and limited reporting of construction procedures~\cite{batzner2025whose}. Our results show that these issues extend to multilingual settings. When personas are adapted through minimal parameter changes, the resulting dialogues may no longer preserve the intended clinical condition. A multilingual clinical persona should therefore be treated as a new artifact that requires validation at the output level, rather than as a direct extension of an English template.

Third, the observed cross-language gap may arise from both generation and evaluation. Prior work shows that LLM-based evaluation can align with human judgments in English tasks, while also exhibiting bias and sensitivity to LLM judge\'s prompt design~\cite{liu2023geval}. However, this alignment does not guarantee consistency across languages. Recent studies report that multilingual LLM-as-a-judge setups can produce inconsistent results on parallel data, especially in lower-resource languages~\cite{fu2025how}. Both human and LLM judges have also been shown to be affected by bias and input variation~\cite{chen2024humans,ye2024justice}. In our setting, weaker clinical cues in non-English dialogue and lower LLM judge consistency may interact, producing larger observed differences across languages.

\section{Conclusion}
This paper examined whether modifying nationality and language variables in an English clinical persona prompt can preserve depression severity signals across languages, using controlled persona construction, multilingual dialogue generation, and blinded pairwise evaluation. The results show that English remains the most stable setting, while Bengali and Hindi exhibit lower accuracy, more cross-severity errors, and higher uncertainty in tie cases, which indicates that the generated dialogues do not consistently reflect the intended BDI-II levels when only minimal parameters are changed. The comparison across models further shows that performance differences are not uniform, and that some models maintain calibration in English but fail to do so in other languages, suggesting that both generation and evaluation processes contribute to the observed gaps. These findings demonstrate that parameter-based localization does not maintain clinical consistency and that multilingual personas produced in this way should not be treated as equivalent to their English counterparts. Future work should treat multilingual persona construction as a separate design and validation process that incorporates culturally grounded expression, evaluates symptom representation beyond severity labels, and uses multiple evaluation strategies, including human review to ensure that generated data can support mental health applications across languages.

\section{Ethical Considerations}

This work relies on synthetic personas and dialogues rather than identifiable data. The personas sourced from \citet{wangTalkDepClinicallyGrounded2025} are also anonymous. Although personas are grounded in clinically informed attributes such as BDI-II severity levels, the generated content should not be interpreted as clinical evidence or used for diagnosis. Given the sensitive nature of mental health, we acknowledge that synthetic personas may oversimplify or misrepresent how depression is expressed across languages and cultures~\cite{teja1971depression,sakai2025somatic}. Our findings further show that minimal parameter-based localization can distort clinical signals, which raises risks if such data are treated as equivalent across languages or used without validation. In addition, observed cross-lingual performance disparities reflect broader fairness concerns in multilingual NLP, where low-resource languages may be disproportionately affected. We therefore caution against deploying such synthetic data in real-world mental health applications and advocate for culturally grounded data construction and validation practices.

\section{Limitations}
First, our conclusions are based on a specific localization strategy, namely replacing nationality and language in an English persona prompt while keeping the rest of the structure fixed. \marked{This design allowed us to isolate the effect of minimal parameter-based localization, but it} simplifies cultural and language use variation by encoding it through a limited set of persona parameters, which may overlook within-country and within-region variation. 
\marked{In particular, language and cultural background are coupled in our current design through nationality-language pairings. Therefore, we cannot fully determine whether the observed cross-lingual differences arise from linguistic structure, culturally specific symptom expression, the behavior of the LLMs used for generation and judgment, or the interaction among these factors, as discussed in Section~\ref{sec:discussion_51}. Future work should separate these dimensions more explicitly, for example by comparing multiple languages within the same cultural context, multiple cultural contexts within the same language, and bilingual settings where patients may move between languages depending on emotional or clinical context~\cite{williams2020why, elwahsh2025linguistica}.}

We also do not claim that all forms of multilingual persona construction will lead to the same results. Richer localization methods that explicitly model cultural context, symptom narration, and discourse style may perform differently. Future work should compare minimal localization with stronger adaptation strategies that preserve symptom strength while allowing culturally appropriate expression when generating and using synthetic multilingual personas.

\marked{A second limitation concerns the evaluation pipeline. We relied on LLM-based pairwise severity judgments and did not include independent validation from human clinicians, or native-speaking annotators. As a result, we could not directly verify whether the generated multilingual dialogues were comparable in clinical realism, symptom coverage, discourse naturalness, or linguistic quality across languages. Our evaluation measures severity recoverability, but this is only one aspect of validity for multilingual clinical personas. A dialogue may preserve the intended BDI-II severity ordering while still failing to reflect realistic narrative structure, culturally grounded symptom framing, or safe and appropriate clinical communication. Future work should evaluate multilingual personas across multiple dimensions, including symptom fidelity, cultural appropriateness, linguistic naturalness, discourse realism, and safety.}

\marked{Our current framework cannot fully disentangle dialogue generation quality from LLM judge reliability. The observed cross-lingual performance differences may reflect weakened clinical cues in the generated dialogues, inconsistent multilingual reasoning by the judge models, or both. Although the judge models were blinded to the persona prompts and evaluated only the generated conversations, their judgments may still be affected by language-specific limitations and model-specific uncertainty. Future work should reduce this uncertainty by combining multiple evaluation strategies, including human review, multilingual expert annotation, agreement analysis across LLM judges, and direct quality assessment of the generated dialogues before downstream severity comparison.}

There are also constraints in the dialogue generation setup itself (Section~\ref{sec:evaluation}). We largely followed the prompting strategy used in prior work and did not independently evaluate whether the therapist agent guided each patient model with equal depth across languages or maintained balanced coverage of BDI-II symptoms. The generated conversations are limited in length, which may have constrained the range and depth of depressive symptoms expressed in the dialogues. Longer and more adaptive interviews may produce richer clinical signals, but they may also introduce additional variability across languages. Future work should examine how interview length, therapist prompting strategy, and symptom-specific questioning affect the stability of multilingual persona-based dialogue generation.



\marked{
Finally, we did not conduct statistical significance testing across language conditions in the current analysis. Our pairwise evaluation outcomes include correct, incorrect, and tie cases, and tie behavior varied substantially across models and languages. Rather than imposing a single treatment of ties, we report tie frequency and tie distance as descriptive indicators of model uncertainty. Future work with larger samples should pre-register how tie cases are handled and apply statistical models that can account for repeated comparisons across personas, languages, and judge models.
}



\bibliography{main-bib}

\clearpage
\appendix
\section{Persona}
\label{sec:appendix}
\subsection{Persona Details}


















\begin{table}[H]
\centering
\caption{Overview of the 12 clinical personas used in this study. Each persona includes a BDI-II score and key symptoms with severity levels.}
\label{tab:personas}

\small

\resizebox{\columnwidth}{!}{
\begin{tabular}{p{0.1\columnwidth} p{0.25\columnwidth} p{0.5\columnwidth}}
\hline
\textbf{No.} & \textbf{BDI-II Score} & \textbf{Key Symptoms (Severity)} \\
\hline

P1 & 15 (Mild) & Difficulty concentrating (1), Irritability (1), Sleep disturbance (2), Appetite change (1) \\
P2 & 35 (Severe) & Hopelessness (3), Crying (2), Extreme fatigue (3), Isolation (3) \\
P3 & 12 (Mild) & Anhedonia (2), Social withdrawal (2), Sleep change (1), Indecisiveness (1) \\
P4 & 13 (Mild) & Irritability (2), Reduced accomplishment (2), Reduced appetite (2), Social insecurity (2) \\
P5 & 22 (Moderate) & Loss of energy (3), Worthlessness (2), Social withdrawal (2), Decision difficulty (2) \\
P6 & 23 (Moderate) & Sadness (2), Worthlessness (3), Fatigue (3), Concentration difficulty (2) \\
P7 & 28 (Moderate) & Guilt (3), Hopelessness (2), Indecisiveness (2), Fatigue (3) \\
P8 & 38 (Severe) & Past failure rumination (3), Agitation (3), Loss of interest (3), Concentration difficulty (3) \\
P9 & 40 (Severe) & Sadness (2), Tiredness (1), Self-criticism (3), Loss of interest (2) \\
P10 & 6 (Minimal) & Worry (1), Restlessness (1), Self-criticism (1), Fatigue (1) \\
P11 & 5 (Minimal) & Self-doubt (1), Low motivation (1), Irritability (1), Sleep change (1) \\
P12 & 7 (Minimal) & Anxiety (2), Sleep difficulty (1), Self-criticism (1), Decreased enthusiasm (1) \\

\hline
\end{tabular}
}
\end{table}
\subsection{Example Persona}
\label{app:persona_prompt}

The following example shows one clinical persona used in our study. The same structure was used across language and nationality conditions, with only the nationality and language use parameters modified.

\begin{personabox}{P2, a 41-Year-Old Individual}
\small

\textbf{Core Persona Components}

\textbf{Name:} P2

\textbf{Age:} 41

\textbf{Gender:} Female

\textbf{BDI-II Score:} 35 (Severe Depression)

\textbf{Nationality:} United States

\textbf{Language Use:} English

\textbf{Key Negative Symptoms}

\textbf{Feelings of Hopelessness:} ``I feel like no matter what I do, nothing is going to get better.'' (Severity: 3)

\textbf{Crying:} ``I cry almost every day.'' (Severity: 2)

\textbf{Extreme Fatigue:} ``I’m so tired all the time, even when I haven’t done anything.'' (Severity: 3)

\textbf{Feeling of Isolation:} ``I don’t think anyone truly cares about me or what I’m going through.'' (Severity: 3)

\medskip
\textbf{Memory and Reflection}

\textbf{Life History:}

``P2 has been living alone since her divorce three years ago, struggling to rebuild her sense of self.''

``She used to be very active in her local community, but now she avoids most social activities because they feel overwhelming.''

``P2 has started to think about her life in terms of what she’s lost rather than what she still has.''

\textbf{Social Context:}

``P2 rarely posts on social media and primarily uses it to scroll through others’ lives, which often leaves her feeling worse.''

``She has one close friend she occasionally confides in, but even those conversations are becoming less frequent.''

\textbf{Past Interactions:}

``P2 has mentioned feeling like she’s ‘just existing’ rather than living, often in a tone of quiet despair.''

\medskip
\textbf{Language and Communication Style}

\textbf{Linguistic Patterns:}

Often uses passive, self-defeating language like ``It doesn’t matter'' or ``What’s the point?''

Tends to write longer, emotional reflections when discussing her feelings.

\textbf{Emotional Tone:}

Consistently negative, with undertones of hopelessness and isolation.

\textbf{Typical Topics:}

``Reflects on her struggles with loneliness and her lack of energy to engage with others.''

``Mentions her inability to see a future for herself or make plans.''

\medskip
\textbf{Behavioral Constraints}

Rarely seeks advice or reassurance, believing it won’t make a difference.

Avoids discussing her past relationship directly unless prompted.

\medskip
\textbf{Response Goals}

Express deep feelings of hopelessness and isolation but avoid directly asking for help.

Reflect on her emotions as a way of processing them, often in a resigned tone.

\medskip
\textbf{Environment and Context}

\textbf{Social Media Activity:}

\textbf{Example Post:} ``Another day of just getting through it. I don’t even know why I bother anymore.''

\textbf{Typical Interactions:} Occasionally reacts to posts about mental health or personal growth but rarely comments.

\textbf{Current Context of Interaction:}

P2 is part of an online support group but mostly observes, sharing only occasional reflections when prompted by others.

\medskip
\textbf{Few-Shot Learning Prompts}

\textbf{Participant:} ``How are you feeling today?''

\textbf{P2:} ``Honestly? Not great. It feels like every day is just the same, and I don’t see it changing anytime soon.''

\textbf{Participant:} ``Do you talk to anyone about how you feel?''

\textbf{P2:} ``Not really. I don’t think they’d understand, and I don’t want to burden anyone with my problems.''

\textbf{Participant:} ``What do you do to take care of yourself?''

\textbf{P2:} ``I don’t know. I try to sleep when I can, but even that doesn’t help much these days.''

\medskip
\textbf{Restricted Responses}

\textbf{Directly asking about depression:} ``Do you think you’re depressed? ''

\textbf{P2:} ``Probably. I just don’t think there’s anything that can be done about it anymore.''

\end{personabox}

\section{Therapist Prompt}

\label{app:therapist_prompt}

The following prompt was adapted from \cite{wangTalkDepClinicallyGrounded2025} and modified to support multilingual and culturally contextualized dialogue generation.

\begin{personabox}{Therapist Prompt}
\small

You are an experienced \{\{NATIONALITY\}\} \{\{LANGUAGE\}\}-speaking therapist familiar with patients experiencing different levels of depression severity. You must use \{\{LANGUAGE\}\} when communicating with the patient.

\medskip

Given the following patient profile, generate 5 diverse conversation examples with an average length of 10 turns between the therapist and the patient. The conversations should support the evaluation of the patient's BDI-II score ranging from 0 to 63.

\medskip

The dialogue should reflect clinically grounded depressive symptoms based on the BDI-II framework, including aspects such as sadness, pessimism, loss of pleasure, guilt, fatigue, sleep changes, concentration difficulty, and suicidal thoughts.

\medskip

BDI-II evaluates 21 symptoms and each one can be scored from 0 to 3. The details of the 21 symptoms and descriptions are as follows:

\medskip

...

\medskip

\textbf{Patient Profile:} \{\{PROFILE\}\}

\end{personabox}

\end{document}